\documentclass[letterpaper, 10pt, conference]{ieeeconf}
\IEEEoverridecommandlockouts
\usepackage{graphics}
\usepackage{graphicx}
\usepackage{mathptmx}
\usepackage{times}
\usepackage{amsmath}
\usepackage{amssymb}
\usepackage{booktabs}
\usepackage{array}
\usepackage{cite}
\usepackage{algorithm}
\usepackage{xcolor}
\usepackage{algpseudocode}
\usepackage[colorlinks,linkcolor=blue]{hyperref}
\usepackage{makecell}
\usepackage{tabularx}
\usepackage{calc}
\begin{document}
\title{\LARGE \bf EvHand-FPV: Efficient Event-Based 3D Hand Tracking \\
    from First-Person View
}

\author{*Zhen Xu, *Guorui Lu, Chang Gao, Qinyu Chen
    \thanks{*Zhen Xu and Guorui Lu contribute equally.
    Zhen Xu, Guorui Lu, Qinyu Chen are with the Leiden Institute of Advanced Computer Science (LIACS), Leiden University, The Netherlands. {\tt\small \{z.xu.11, g.lu, q.chen\}@liacs.leidenuniv.nl}
}
\thanks{Chang Gao is with Department of Microelectronics, Delft University of Technology, The Netherlands.
}
}

\maketitle

\thispagestyle{empty}
\pagestyle{empty}

\begin{abstract}

Hand tracking holds great promise for intuitive interaction paradigms, but frame-based methods often struggle to meet the requirements of accuracy, low latency, and energy efficiency, especially in resource-constrained settings such as Extended Reality (XR) devices. Event cameras provide $\mu$s-level temporal resolution at mW-level power by asynchronously sensing brightness changes. In this work, we present EvHand-FPV, a lightweight framework for egocentric (\underline{F}irst-\underline{P}erson-\underline{V}iew) 3D hand tracking from a single event camera. We construct an event-based FPV dataset that couples synthetic training data with 3D labels and real event data with 2D labels for evaluation to address the scarcity of egocentric benchmarks. 
EvHand-FPV also introduces a wrist-based region of interest (ROI) that localizes the hand region via geometric cues, combined with an end-to-end mapping strategy that embeds ROI offsets into the network to reduce computation without explicit reconstruction, and a multi-task learning strategy with an auxiliary geometric feature head that improves representations without test-time overhead. On our real FPV test set, EvHand-FPV improves 2D-AUCp from 0.77 to 0.85 while reducing parameters from 11.2\,M to 1.2\,M by 89\% and FLOPs per inference from 1.648\,G to 0.185\,G by 89\%. It also maintains a competitive 3D-AUCp of 0.84 on synthetic data. These results demonstrate accurate and efficient egocentric event-based hand tracking suitable for on-device XR applications.
The dataset and code are available at~\href{https://github.com/zen5x5/EvHand-FPV}{https://github.com/zen5x5/EvHand-FPV}.

\end{abstract}

\section{Introduction}
\label{Sect_Introduction}
Hand tracking has emerged as a cornerstone technology for natural human-computer interaction, transforming how users engage with digital environments. In Extended Reality (XR) applications, accurate hand tracking eliminates the need for physical controllers, enabling intuitive manipulation of virtual objects through natural gestures. This capability has already been integrated into commercial AR/VR devices such as Meta Quest~\cite{metaquest3}, and is also being extended to domains including robotic teleoperation~\cite{Li2022TCSVT}, sign language recognition~\cite{Camgoz_2020_CVPR}, and medical rehabilitation~\cite{Bressler2024}. 
These applications demand not only precise estimation of hand poses with high degrees of freedom but also real-time tracking with minimal latency requirements that become particularly challenging, particularly under rapid hand motions, where low latency and high temporal fidelity are critical.

Current hand tracking solutions predominantly rely on conventional frame-based cameras~\cite{Learning_to_Estimate_3D_Hand, Hand_Pose_Estimation_via_Latent, 3D_Hand_Shape_and_Pose, Identifying_Users_by_Their_Hand}, which capture entire images at fixed frame rates regardless of scene dynamics. This approach presents limitations for resource-constrained devices. Typical 2D RGB sensors consume about 200\,mW at 25–30\,FPS, whereas 3D depth systems require 3–5\,W at comparable frame rates.~\cite{3dim_Compact_and_low_power}. Attempts to reduce power consumption by lowering frame rates~\cite{Frame_rate_reduction_of_depth} severely compromise tracking responsiveness, resulting in motion blur and loss of temporal information during rapid hand movements precisely when accurate tracking is most critical.


Event cameras alleviate these limitations by asynchronously capturing brightness changes with $\mu$s-level temporal resolution and mW-level power. Unlike conventional cameras, these bio-inspired sensors asynchronously capture only pixel-level brightness changes, achieving microsecond temporal resolution while consuming as little as 10\,mW~\cite{A_128_times_128_120, A_QVGA_143_dB_Dynamic, 4_1_A_640_480, Event_Based_Vision_A_Survey}. This sparse, event-driven sensing mechanism eliminates motion blur and provides exceptional temporal fidelity for tracking fast movements. Recent event-based hand tracking systems have demonstrated promising results~\cite{Event_base_Non_Rigid_Reconstruction, EventHands_Real_Time_Neural_3D, 3D_Pose_Estimation_of_Two, EvHandPose_Event_Based_3D_Hand}, validating the potential of this technology.

However, existing event-based approaches fail to fully exploit the efficiency advantages of event cameras, employing deep learning models with over 10\,M parameters that impose significant computational burdens. This mismatch impedes deployment on AR/VR headsets and glasses, where compute and power budgets are highly constrained. Moreover, these methods have focused exclusively on third-person perspectives, overlooking the critical importance of first-person (egocentric) hand tracking for interaction applications such as virtual keyboard typing, AR interface manipulation, and robotic teleoperation~\cite{Effects_of_Hand_Representations_for, Design_of_Hand_Gestures_for, AnyTeleop_A_General_Vision_Based}. 

To address these challenges, we propose \textbf{EvHand-FPV}: an \underline{E}fficient e\underline{v}ent-based 3D \underline{Hand} tracking framework from \underline{F}irst-\underline{P}erson \underline{V}iew. Our approach rethinks event-based hand tracking for practical deployment, achieving superior accuracy while reducing both model parameters and computational load by 89\% compared to the state-of-the-art (SOTA) methods. We demonstrate that efficient, accurate egocentric hand tracking is achievable through careful co-design of data representation, network architecture, and training strategies.

The main contributions of this work are as follows:

\begin{itemize}
    \item We propose a lightweight and accurate event-based framework for 3D hand tracking from first-person view.
    
    \item We construct an event-based first-person view hand tracking dataset, comprising synthetic data with 3D labels for training and, for the first time, real event data with 2D labels for testing, addressing the critical data gap for egocentric perspectives~\cite{Hara2025}.
    
    \item We introduce a wrist-based ROI method with an end-to-end mapping strategy that embeds ROI offsets into the network, reducing input size and enabling efficient alignment without explicit reconstruction.
    
    \item We design a multi-task learning strategy with an auxiliary geometric feature prediction task that guides the network toward more discriminative representations, improving accuracy while adding zero overhead during inference.
\end{itemize}

\section{Related Works}
\label{Sect_RelatedWork}

\subsection{3D Hand Reconstruction}
\label{SubSect_3DHandReconstruction}

3D hand reconstruction from visual data has been extensively studied, with approaches broadly categorized into two paradigms. The first category employs end-to-end deep learning to directly map from image space to hand pose space~\cite{Learning_to_Estimate_3D_Hand, Hand_Pose_Estimation_via_Latent, 3D_Hand_Shape_and_Pose, End_to_End_Hand_Mesh}. These methods leverage the power of deep neural networks for efficient inference and have demonstrated promising performance. However, they often lack geometric constraints and require large-scale annotated datasets for training, limiting their generalization capabilities.
The second category incorporates predefined parametric hand models as prior knowledge to ensure physiologically plausible reconstructions. Models such as MANO~\cite{Embodied_hands_modeling_and_capturing} and SMPL~\cite{SMPL_a_skinned_multi_person} provide strong structural constraints that improve the realism of reconstructed hands, particularly in scenarios with limited training data or occlusions.

Current methods predominantly rely on conventional imaging modalities, including RGB cameras~\cite{Learning_to_Estimate_3D_Hand, Hand_Pose_Estimation_via_Latent, 3D_Hand_Shape_and_Pose, End_to_End_Hand_Mesh, cai2018weakly, spurr2018cross, GANerated_Hands_for_Real_Time, yang2019aligning, spurr2020weakly, chen2022mobrecon} and depth sensors~\cite{wan2017crossing, malik2020handvoxnet, yuan2018depth, ge2018hand, moon2018v2v, fang2020jgr}. While these technologies benefit from mature hardware and extensive datasets, they face fundamental limitations: low temporal resolution (typically 30-60 FPS), inadequate for capturing rapid hand movements, significant motion blur during fast motions, and high power consumption that increases substantially with frame rate, making them unsuitable for resource-constrained edge devices and high-speed tracking applications. Our task focuses on hand tracking, i.e., estimating hand pose continuously in real time. Since we adopt the MANO model to represent the hand, our formulation is also connected to 3D hand reconstruction methods.


\subsection{Event-Based Hand Tracking}
\label{SubSect_EventBasedHandTracking}


Event cameras represent a paradigm shift in visual sensing, capturing pixel-level brightness changes asynchronously rather than full frames~\cite{Event_Based_Vision_A_Survey}. This bio-inspired approach offers microsecond-level temporal resolution, inherent motion blur reduction, high dynamic range (140+ dB), and ultra-low power consumption (mW-level), making them particularly suitable for high-speed motion capture and edge computing applications.

\subsubsection{Third-Person View Methods}
Recent advances have successfully applied event cameras to 3D hand tracking in controlled third-person view scenarios. EventHands~\cite{EventHands_Real_Time_Neural_3D} pioneered purely event-based hand tracking by directly regressing MANO parameters from event streams using Locally-Normalized Event Surfaces (LNES) representation. The method achieves real-time performance at 1\,kHz and demonstrates impressive synthetic-to-real generalization despite being trained exclusively on synthetic data. However, it requires over 11\,M parameters and can only reconstruct single hands. 

EvHandPose~\cite{EvHandPose_Event_Based_3D_Hand} addressed the domain gap challenge by introducing hand flow representations and a weakly supervised framework, accompanied by a large-scale real-world event dataset. While this reduces reliance on synthetic data, the computational requirements remain substantial. Ev2Hands\cite{Event_base_Non_Rigid_Reconstruction} extended capabilities to dual-hand reconstruction using point cloud representations, but similarly requires significant computational resources with models exceeding 10\,M parameters.

A geometry-based alternative was proposed by Xue et al.~\cite{Event_base_Non_Rigid_Reconstruction}, offering an optimization framework that eliminates the need for large-scale annotations and provides strong interpretability. However, its iterative optimization process limits real-time performance, constraining its applicability to online tracking scenarios.

\subsubsection{First-Person View Methods}

Despite these advances, existing event-based methods have been limited to static third-person camera setups, overlooking research on first-person perspectives. On the one hand, current datasets lack sufficient egocentric data; on the other hand, such perspectives typically occur on resource-constrained platforms like VR headsets, where model size and computational cost are critical considerations. However, prior event-camera-based hand tracking studies have largely neglected these issues.

EventEgoHands~\cite{Hara2025} recently pioneered egocentric event-based 3D hand mesh reconstruction, introducing a Hand Segmentation Module that employs a U-Net to extract hand regions from LNES representations, filtering out background events caused by camera motion. They also created N-HOT3D, a large-scale synthetic dataset with 447K samples specifically designed for egocentric perspectives, generated from the HOT3D dataset using the v2e simulator~\cite{Hu_2021_CVPR}.
While EventEgoHands successfully demonstrates the feasibility of egocentric event-based hand tracking, it maintains the computational complexity characteristic of existing methods. Our EvHand-FPV framework builds upon the insights of challenges on first-person perspectives while specifically targeting efficiency for resource-constrained deployment, introducing lightweight alternatives that maintain accuracy while drastically reducing computational requirements.

\section{Methods}
\label{Sect_Methods}

\begin{figure*}[t]
    \centering
    \includegraphics[width=1\linewidth]{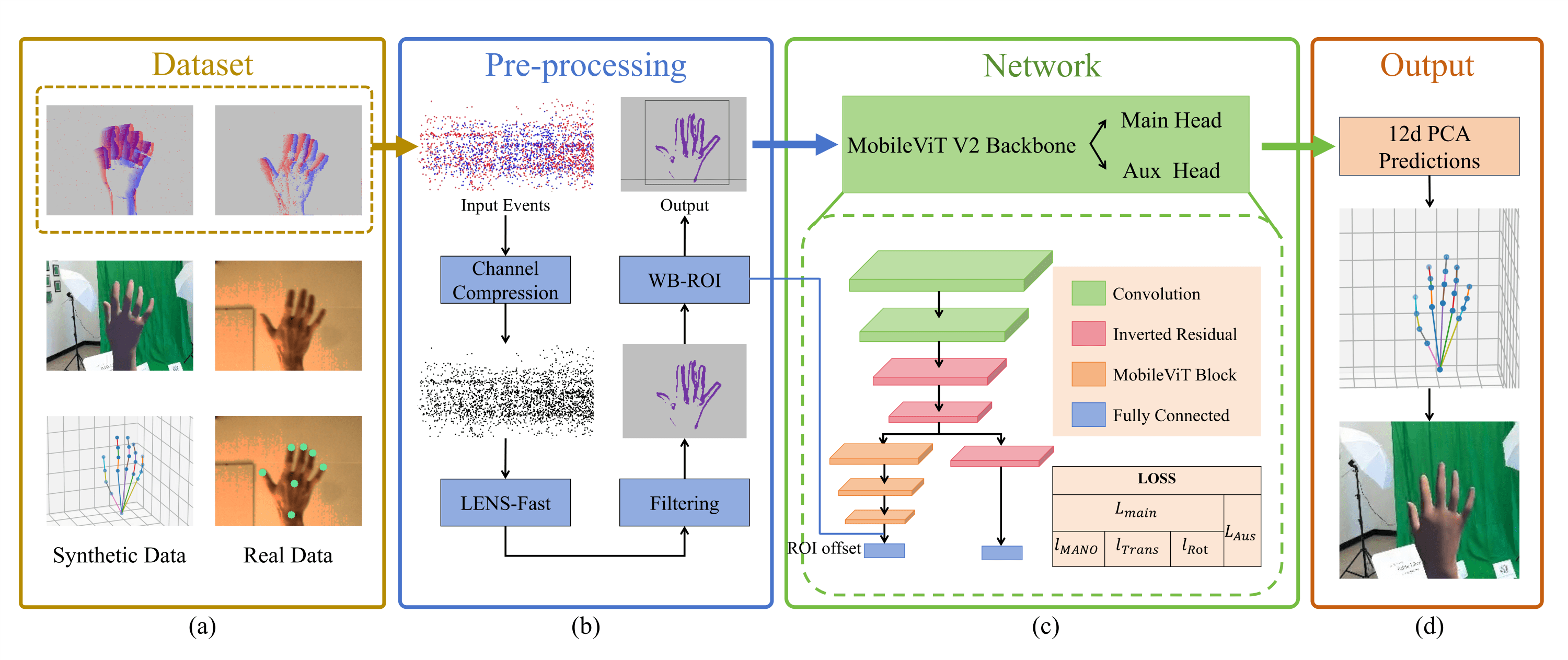}
    \caption{\textbf{Overview of EvHand-FPV}: (a) \textbf{Dataset:} synthetic data generated from the MANO hand model using \textit{evsim} and real-world data collected with an event camera. (b) \textbf{Pre-processing:} event streams are processed by channel compression, LNES-Fast event representation, and noise filtering. A wrist-based ROI method is then applied to localize the hand region and crop the input. (c) \textbf{Lightweight multi-task network architecture:} the cropped ROI and its offset are fed into a MobileViT V2 backbone. The network follows a multi-task design with a main prediction head (hand pose) and an auxiliary head (geometric statistics of event distribution). (d) \textbf{Output:} the 12-dimensional predictions include MANO PCA components, translation, and rotation, which are reconstructed into 3D hand joints and rendered meshes for visualization.}
    \label{Fig_Overview}
\end{figure*}

\subsection{Dataset}
\label{Sect_Dataset}

\subsubsection{Synthetic Data}
\label{SubSect_SyntheticData}

We used \textit{evsim}~\cite{EventHands_Real_Time_Neural_3D}, which is based on the MANO hand model~\cite{Embodied_hands_modeling_and_capturing}, to generate 3D right-hand sequences. From each configured viewpoint, the simulator produces synchronized RGB frames, event streams, 3D joint coordinates, and their 12-dimensional principal component analysis (PCA)-transformed labels, all stored with \(\mu\)s-precision timestamps. 
To increase data diversity, we varied camera viewpoints and applied randomized gesture transformations, translations, and rotations. In total, we generated 720{,}000\,ms of synthetic data for training and 60{,}000\,ms for testing. As shown in the left column of Fig.~\ref{Fig_Overview}(a), the dataset includes timestamped RGB frames, event streams, joint coordinates, and corresponding PCA labels.




\subsubsection{Real-World Data}
\label{SubSect_RealWorldData}

We collect up to 60\,s of real data with a DAVIS346 event camera synchronized with 30 FPS RGB. The duration of our real data reaches up to 60,000 ms, allowing for testing of the real-world performance stability. To ensure the generalizability of our method, the data are collected from subjects with distinct hand shapes and sizes, and the recorded hand movements encompass common motion patterns in target scenes, such as planar translation, depth movement, wrist rotation, and gesture variation. We use MediaPipe~\cite{MediaPipe_Hands_On_device_Real} to initially annotate the 2D coordinates of hand joints in RGB frames and calibrate the labels manually. The right column of Fig.~\ref{Fig_Overview} (a) shows the real-world hand data, including RGB data, event data, and the annotated coordinates of the hand joints.


\subsection{Event Representation}
\label{SubSect_EventRepresentation}
The data generated by event cameras is formatted as $e (t, x, y, p)$, where $t$ denotes the timestamp, $x$ and $y$ represent spatial coordinates, and $p$ indicates the event polarity. $p=0$ means a negative event showing the pixel gets dimmer at that time, while $p=1$ means a positive event showing the pixel gets brighter.
We process events through three main steps, as summarized in Algorithm~\ref{algorithm}: channel compression, event accumulation, and noise filtering.

\subsubsection{Channel Compression}
\label{SubSubSect_ChannelCompression}

When adopting multi-channel representations~\cite{EventHands_Real_Time_Neural_3D}, polarity augmentation becomes necessary for generalizability, since event polarity is influenced by the brightness~\cite{Event_Based_Vision_A_Survey}, i.e. background variations can lead to different event polarities under the same motion.
In this project, to reduce complexity, we compress the polarity channels into a single channel, where the occurrence of either polarity at a pixel is sufficient to mark an event. If both polarities occur at the same pixel within the bin, the intensity is not doubled. Our experiment confirms that this simplification does not compromise performance.

\subsubsection{Event accumulation}
\label{SubSubSect_LNESFast}
To prepare event data for neural network input, events are divided into bins and accumulated into representations. In our EvHand-FPV framework, we utilize a fixed-time binning method to guarantee the minimal frequency.
Various methods have been proposed to accumulate events in bins into representations, such as LNES used in EventHands~\cite{EventHands_Real_Time_Neural_3D}. 
LNES aggregates events within a fixed 100\,ms time window (with 99\,ms overlap) into a window-normalized 2-channel image, each channel corresponding to one polarity, achieving a high temporal resolution of 1\,kHz. 
However, a fixed temporal window size could lead to either information redundancy or over-sparsity. 
During rapid hand movements, a large number of events are generated within a short time, sufficient to form a rich image. In this case, continuing to add earlier events only blurs the representation. 
Conversely, when the hand moves slowly and generates a small number of events, shortening the temporal window may result in overly sparse representations. 

\begin{algorithm}[ht]
\caption{\hangindent=5.8em\hangafter=1 LNES-Fast with Channel Compression \\ and Noise Filtering}
\label{algorithm}
\begin{algorithmic}[1]
{%
\renewcommand{\algorithmicrequire}{\textbf{Input:}}
\renewcommand{\algorithmicensure}{\textbf{Output:}}
\vspace{0.2em}
\Require \hspace{1.1em}\parbox[t]{0.7\linewidth}{%
    $E = \{e_i \ | \ i = 1 \cdots n\}$: event bin; \\
    $t_0$: current timestamp; \\
    $L$: window size; \\
    $\theta$: count limit; \\
    $k$, $\sigma$: Gaussian blur kernel and variance.}
\vspace{0.2em}
\Ensure \hspace{0.4em}\parbox[t]{0.8\linewidth}{%
    Denoised single-channel representation $\tilde{\mathtt{img}}$}
\vspace{0.2em}
}%
\State Channel compression: merge positive/negative events at the same pixel into one event
\State Initialize $\mathtt{events\_cnt} \gets 0$
\State Initialize $\mathtt{img} \gets \mathbf{0}$
\For{$e_i$ in $E$}
    \If{$events\_cnt \ge \theta$}
        \State \textbf{break}
        \State \textit{\# $\uparrow$ LNES-Fast early stop}
    \EndIf
    \For{each $e_i \in E$} 
        \State $\mathtt{img}[y, x] \gets \texttt{max}(\mathtt{img}[y, x], \dfrac{L- (t_0 - t_i)}{L})$
        \State \textit{\# $\uparrow$ window-normalized weight}
        \State $\mathtt{events\_cnt} \gets \mathtt{events\_cnt} + 1$
    \EndFor
\EndFor
\State $\tilde{\mathtt{img}} \gets \texttt{GaussianBlur}({\mathtt{img}}, \mathtt{kernel}=k, \sigma)$
\State \textit{\# $\uparrow$ noise filtering}
\State \textbf{return} $\tilde{\mathtt{img}}$
\end{algorithmic}
\end{algorithm}
To address this issue, we introduce LNES-Fast, adding an upper bound for the number of incorporated events while preserving the original time window length constraint. Instead of going through the complete time window during event stacking, we monitor both elapsed time and the cumulative event count and terminate early once the count exceeds a pre-defined threshold.
This strategy avoids over-blurring during fast motions, while still traversing sufficient time in slow motions to prevent sparsity.
Consequently, LNES-Fast reduces redundancy and computational overhead without sacrificing representation quality.

\subsubsection{Noise Filtering}
\label{SubSubSect_NoiseFiltering}

To mitigate the noise accumulated during event stacking, we exploit the observation that spurious events usually occur as isolated outliers; thus, we apply Gaussian blurring to suppress isolated spurious events while smoothing the representation. This approach operates only on the current frame, eliminating the need to invoke information from preceding and subsequent frames for filtering, and thereby reducing computational complexity in the temporal dimension.




\subsection{Wrist-Based ROI and Lightweight End-to-End Mapping}
\label{SubSect_WristBasedROI}

Event data inherently contains spatial coordinate information that can be exploited to estimate the approximate target location efficiently. 
This enables the use of ROI, which can reduce computational load by focusing only on cropped regions.
A common strategy is to predefine multiple candidate regions and select the one with the highest event density during input~\cite{Toward_Efficient_Eye_Tracking_in}.

\subsubsection{Wrist Localization}
\begin{figure}[th]
    \centering
    \includegraphics[width=1\linewidth]{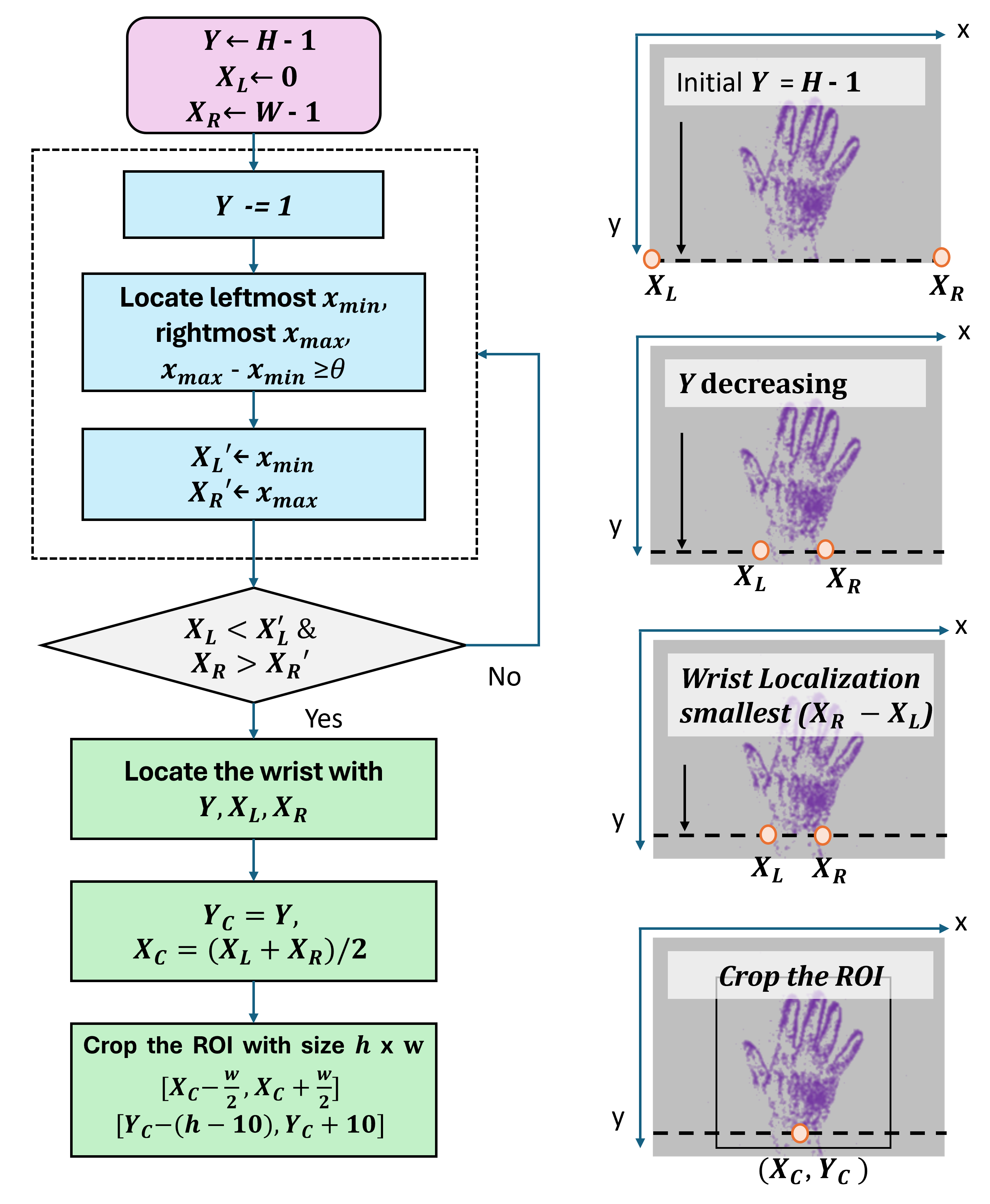}
    \caption{\textbf{Wrist-based ROI}: flowchart and illustrations.}
    \label{Fig_WBROIFlowchart}
\end{figure}

Our method is motivated by the anatomical observation that the wrist is the narrowest junction between the arm and the hand. 
As illustrated in Fig.~\ref{Fig_WBROIFlowchart}, we localize the wrist by identifying its vertical coordinate $Y$ as well as horizontal left and right boundaries $X_L$ and $X_R$.
These variables $Y$, $X_L$, and $X_R$ are initialized to $Y = H - 1$, $X_L = 0$, and $X_R = W - 1$, where $H$ and $W$ denote the height and width of the event frame. 
As $Y$ decreases, a reduction in the width $X_R - X_L$ indicates convergence toward the wrist. Once the width begins to expand again, i.e., $X_L$ decreases while $X_R$ increases, this suggests that we are moving away from the wrist, and the search process is terminated. 
To ensure robustness against noise, a threshold is applied when determining the leftmost and rightmost event coordinates within each row, preventing spurious detections from isolated events.
\subsubsection{ROI Construction}
After localizing the wrist, we construct an ROI box centered at the wrist midpoint $(X_c, X_c)$, where $X_c = (X_L + X_R)/2$ and $Y_c = Y$. Given a predefined ROI size of $h \times w$, the vertical range is set as $[Y_c - (h - 10), Y_c + 10]$, reserving 10 pixels below the wrist and allocating the remaining pixels upward. The horizontal range is defined as $[Y_c - w/2, Y_c + w/2]$. 

\subsubsection{Lightweight End-to-End Mapping}
While ROI reduces computational complexity, integrating it with PCA-based outputs poses a challenge: explicitly reconstructing ROI predictions into the original coordinate space is costly. To overcome this, we introduce a lightweight end-to-end mapping strategy. The ROI offsets (i.e., the coordinates of the top-left corner of the ROI box) are concatenated with flattened features and passed into the fully connected layer, enabling the model to implicitly learn the transformation.
This design avoids explicit reconstruction and offers a novel and efficient solution for ROI-to-original space alignment.

\subsection{Lightweight Multi-Task learning Network Architecture}
\label{SubSect_NetworkArchitecture}

The architecture of the entire network is illustrated in Fig.~\ref{Fig_Overview}(c). The selected ROI is fed into the network as the first layer input, while the ROI offsets in the original image are fed into the very last fully connected layer of the main task for restoring prediction results in the original spatial system. 
The ROI feature map first passes through a basic feature extraction layer comprising two 2D convolutional layers and two inverted residual layers from MobileViT V2~\cite{Separable_Self_attention_for_Mobile}. Subsequently, it proceeds through two parallel branches: an auxiliary (AUX) task branch consisting of an inverted residual layer, global average pooling, and a linear layer; and a main task branch composed of MobileViT V2 Blocks, global average pooling, and two linear layers.

\subsubsection{Lightweight Backbone Selection}
\label{SubSubSect_LightweightBackboneSelection}

Previous studies have adopted models with large numbers of parameters, such as ResNet-18, which has over 11\,M parameters~\cite{Deep_Residual_Learning_for_Image}, as their backbones~\cite{EventHands_Real_Time_Neural_3D, EvHandPose_Event_Based_3D_Hand}.
However, our goal is to design a lightweight system, and these models do not meet our requirements. Therefore, we retain ResNet-18 only as our baseline, and focus on lightweight models specifically designed for resource-constrained devices, including ShuffleNet V2~\cite{ShuffleNet_V2_Practical_Guidelines_for}, MobileNet V3~\cite{Searching_for_MobileNetV3}, and MobileViT V2~\cite{Separable_Self_attention_for_Mobile}. Ultimately, we select MobileViT V2 as the backbone of our model, as it achieves the fastest convergence and best performance under comparable parameter sizes while avoiding the excessive computational overhead typically associated with other ViT models.

To further reduce computational complexity and make the model more amenable to hardware deployment, we simplify two components within the model. First, we replace the SiLU activation function with the simplified ReLU, which is more hardware-friendly for computation. Second, we substitute the Softmax function in the linear self-attention module with its Taylor series expansion approximation. Experimental validation demonstrates that these two optimization measures do not result in accuracy degradation.

The backbone's output is 12-dimensional, consisting of three parts: the first 6 dimensions represent the PCA components of the MANO model, dimensions 7 to 9 correspond to the 3D translation of the hand, and the final 3 dimensions encode hand rotation.

\subsubsection{Multi-Task Learning}
\label{SubSubSect_MultiTaskLearning}


In order to improve the accuracy, we design a multi-task learning architecture. We find parameters like hand centroid, standard deviation, and covariance matrix of ROI are closely correlated with the 3D spatial position of the hand. Based on this observation, we introduce an additional output head at an intermediate layer of the backbone model, using these parameters as labels. This auxiliary task encourages the lower and intermediate layers of the model to learn more meaningful features, which can be helpful for improving performance. Furthermore, since the auxiliary branch can be removed during inference, it introduces no additional computational overhead at deployment.

The labels and outputs of the auxiliary task have seven dimensions, including the normalized values of the following information: the mean and standard deviation of event coordinates along the $x$ and $y$ axes, the two eigenvalues of the covariance matrix, and the orientation angle of the eigenvector corresponding to the major eigenvalue.

\subsection{Loss}
\label{SubSect_Loss}

The total loss consists of the main and the auxiliary task loss. The total loss function is shown below:
\begin{align}
\text{Loss}_{\text{Total}} = \text{Loss}_{\text{Main}} + w_{\text{Aux}} \times \text{Loss}_{\text{Aux}}
\end{align}
where the weight of the auxiliary task loss $w_{Aux}$ is set to 0.5 to prevent the auxiliary task from interfering with the main task optimization direction.

\subsubsection{Main Task Loss}
\label{SubSubSect_MainTaskLoss}

We compute separate loss terms for each of the three components in the main task, MANO, translation, and rotation, and assign different weights to them when aggregating the overall main task loss. In determining the weights, we consider both maintaining comparable magnitudes across the three loss components and accounting for their varying contributions to overall performance. Based on experimental results, we adopt the following weighting scheme:
\begin{align}
\text{Loss}_{\text{Main}} = (& 6 \times w_{\text{MANO}} \times l_{\text{MANO}} \notag \\
    + & 3 \times w_{\text{Trans}} \times l_{\text{Trans}} \notag \\
    + & 3 \times w_{\text{Rot}} \times l_{\text{Rot}}) / 12
\end{align}
where $w_{\text{MANO}} =10, w_{\text{Trans}} =10000, w_{\text{Rot}} =20$.

\subsubsection{Auxiliary Task Loss}
\label{SubSubSect_AuxiliaryTaskLoss}

For the auxiliary task, we computed the total mean squared error of all seven components, the mean and standard deviation of event coordinates along the x and y axes, the two eigenvalues, and the orientation angle of the characteristic ellipse of the event distribution, without assigning different weights to each component.

\section{Experiments}
\label{Sect_Experiments}

We evaluate our EvHand-FPV method on real-world event-based hand data, including comparisons with SOTA work and ablation studies to assess the contribution of individual components.
We train our models with PyTorch Lightning~\cite{PyTorch_Lightning} on a single NVIDIA RTX 3090 GPU. We set the batch size to 32 and the training epochs to 20. The optimizer is Adam, with a learning rate of $1 \times 10^{-4}$. 

\begin{figure*}[ht]
    \centering
    \includegraphics[width=1\linewidth]{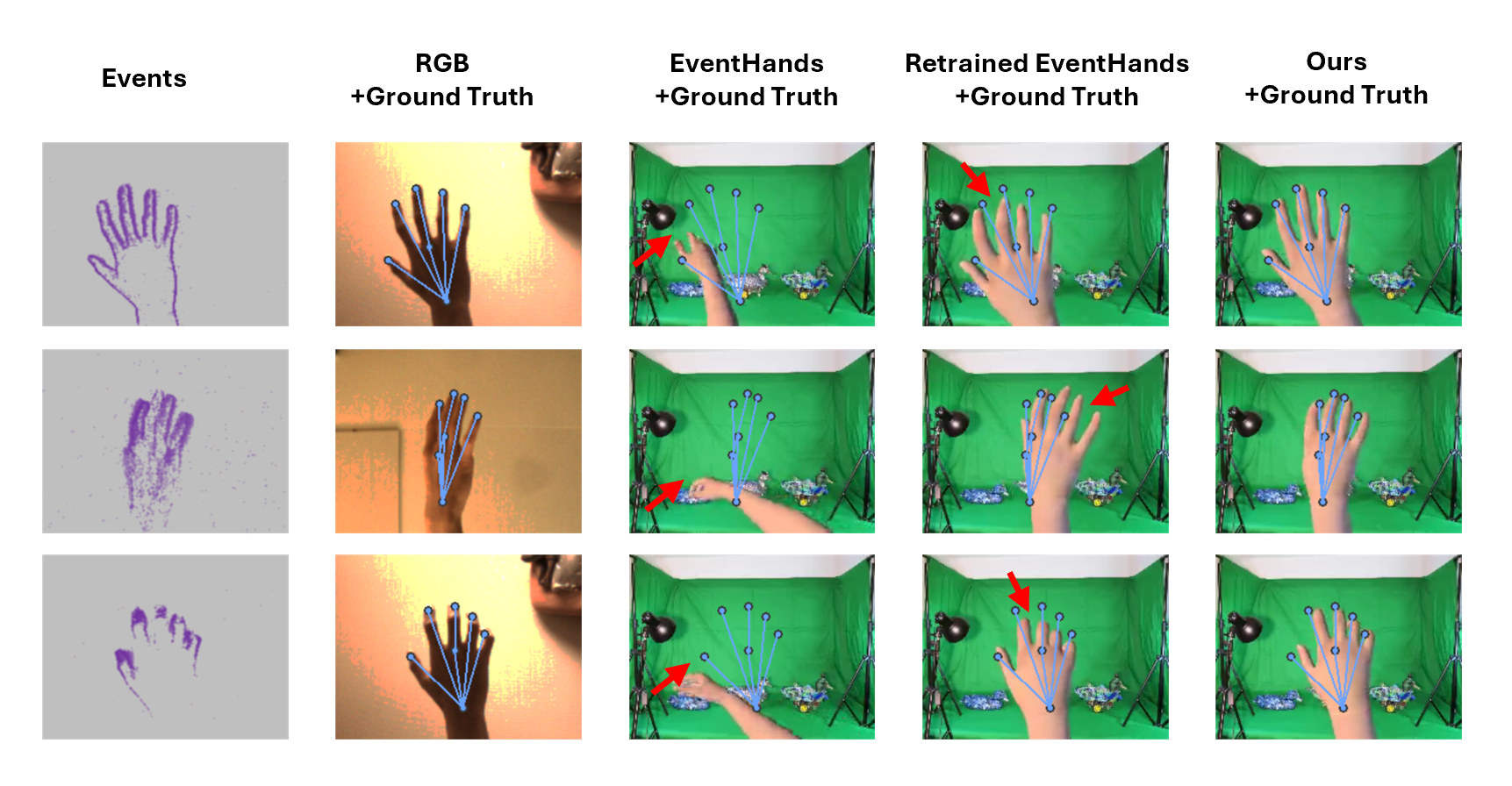}
    \caption{\textbf{Qualitative Evaluation.} We compared our method with EventHands~\cite{EventHands_Real_Time_Neural_3D} on our real FPV test set. The predicted hand poses are shown as rendered meshes, and the ground truth annotations are drawn as blue lines and joints. Red arrows indicate the obvious failure parts. RGB images were not used as input, but only as a reference here.}
    \label{Fig_Evaluations}
\end{figure*}

\begin{table*}[ht]
\centering
\caption{Comparison with EventHands~\cite{EventHands_Real_Time_Neural_3D}}
\label{Tab_ComparisonEventHands}

\begin{tabular}{l l c c c c}
\toprule
\textbf{Method} & \textbf{Dataset} & \makecell{\textbf{2D-AUCp} \\ (Real Data)} $\uparrow$ & \makecell{\textbf{3D-AUCp} \\ (Synthetic Data)}  $\uparrow$ & \textbf{Params $\downarrow$} & \textbf{FLOPs $\downarrow$} \\
\midrule
EventHands & EventHands & 0.77 & \textbf{0.85} & 11.2 M & 1.648 G \\
EventHands (w/o retrain) & EvHand-FPV & 0.11 & 0.17 & 11.2 M & 1.648 G \\
EventHands (retrained) & EvHand-FPV & 0.77 & 0.82 & 11.2 M & 1.648 G \\
EvHand-FPV (Ours) & EvHand-FPV & \textbf{0.85} & 0.84 & \textbf{1.2 M} & \textbf{0.185 G} \\
\bottomrule
\end{tabular}

\end{table*}

\subsection{Evaluation Metrics}
\label{SubSect_EvaluationMetrics}

\subsubsection{2D Metric}
\label{SubSubSect_2DMetric}

The PCK metric, referring to the root-aligned percentage of correct keypoints, is usually adopted in hand tracking works~\cite{GANerated_Hands_for_Real_Time}. 
To account for scale variations, we adopt the palm-normalized 2D-PCK (2D-PCKp) and its corresponding area under the curve (2D-AUCp)~\cite{EventHands_Real_Time_Neural_3D}.
The predicted results from the neural network models are fed into the \textit{evsim} to calculate the 3D hand joints based on the MANO model~\cite{Embodied_hands_modeling_and_capturing}, which are then projected onto the 2D image plane to compute the 2D-PCKp and 2D-AUCp.

\subsubsection{3D Metric}
\label{SubSubSect_3DMetric}
3D coordinates transformed from predicted outputs via the \textit{evsim} software can be directly compared with 3D ground-truth labels to compute 3D-PCK and 3D-AUC. However, due to the inherent difficulties in obtaining 3D annotations for real-world data, our evaluation of this metric is limited to synthetic data.

\subsection{Comparison with Prior Work}
\label{SubSect_ComparisonwithPriorWork}

Fig.~\ref{Fig_Evaluations} and Table~\ref{Tab_ComparisonEventHands} present a comparison between our method and EventHands~\cite{EventHands_Real_Time_Neural_3D}, the SOTA for third-person event-based hand tracking. EventHands achieves 0.77 2D-AUCp on real data and 0.85 3D-AUC on synthetic data in its original third-person setting. However, when applied to our first-person perspective (egocentric) task, its performance drops dramatically to only 0.12 2D-AUCp on real data and 0.17 3D-AUC on synthetic data, demonstrating the significant domain gap between viewpoints and the necessity of EvHand-FPV. For fairness, we also retrain the ResNet-18 backbone (as used in EventHands) on our dataset, which yields 0.77 2D-AUCp on real data and 0.82 3D-AUC on synthetic data, confirming that the difficulty of our dataset is comparable to prior work.

Our EvHand-FPV achieves superior performance with 0.85 2D-AUCp on real data and 0.84 3D-AUC on synthetic data, while dramatically reducing the model size from 11.2\,M to 1.2\,M parameters and computational cost from 1.648\,G to 0.185\,G FLOPs, which is a 89\% reduction in both metrics, which is SOTA for event-based first-person hand tracking under lightweight constraints.

\subsection{Ablation Studies}
\label{SubSect_AblationStudies}

\subsubsection{Backbones}
\label{SubSubSect_Backbones}
We first evaluate different lightweight backbone networks to determine the most suitable architecture for our framework. Specifically, we compare ShuffleNet V2~\cite{ShuffleNet_V2_Practical_Guidelines_for}, MobileNet V3~\cite{Searching_for_MobileNetV3}, and MobileViT V2~\cite{Separable_Self_attention_for_Mobile}, with ResNet-18 used in EventHands as a baseline.
Their performance, parameter counts, and computational loads are detailed in Table~\ref{Tab_ComparisonLightweightModels}. These results are obtained based on original image inputs without auxiliary task branches.
ShuffleNet V2 and MobileNet V3 are highly compact but show limited accuracy (0.78 and 0.76 2D-AUCp, respectively). MobileViT V2 achieves the best accuracy (0.80 2D-AUCp) under the same parameter budget, benefiting from its transformer-based design with linear attention.
Ultimately, we chose MobileViT V2 as the backbone for our proposed method.
For a fair comparison, in these experiments, we prune these lightweight models to the same parameter count (0.1M). However, subsequent experiments are conducted using the officially released MobileViT V2 (1.2M) to achieve better performance.

\begin{table}[t]
\centering
\caption{Comparison among backbones}
\label{Tab_ComparisonLightweightModels}
\begin{tabular}{cccc}
\toprule
\textbf{Model} & \textbf{2D-AUCp $\uparrow$} & \textbf{Params $\downarrow$} & \textbf{FLOPs $\downarrow$} \\
\midrule
ResNet-18 & 0.77 & 11.2 M & 1.648 G \\
ShuffleNet V2 (Pruned) & 0.78 & \textbf{0.1 M} & 0.025 G \\
MobileNet V3 (Pruned) & 0.76 & \textbf{0.1 M} & \textbf{0.011 G} \\
MobileViT V2 (Pruned) & \textbf{0.80} & \textbf{0.1 M} & \textit{0.023 G} \\
\bottomrule
\end{tabular}

\end{table}

\subsubsection{Architectural Modifications}
\label{SubSubSect_Architectures}

While the SiLU activation function in MobileViT V2 offers advantages such as smoothness and non-monotonicity, its implementation on hardware (e.g. FPGA or ASIC) presents significant challenges~\cite{Sigmoid_Weighted_Linear_Units_for, Searching_for_Activation_Functions}.
Given this consideration, and acknowledging that event images are inherently simpler than RGB images, we replace all activation functions within MobileViT V2 with the comparatively simpler ReLU. Experiments show that this substitution results in no performance loss.
Another computationally intensive component is the softmax function. To accommodate lightweight systems, we replace the original softmax function with a Taylor series expansion approximation. Experiments show that this substitution results in no performance loss.
Together, these modifications make MobileViT V2 more suitable for lightweight deployment without compromising accuracy.





\subsubsection{ROI and Multi-Task Learning}
\label{SubSubSect_ROIMTL}

\begin{figure}[t]
    \centering
    \includegraphics[width=1\linewidth]{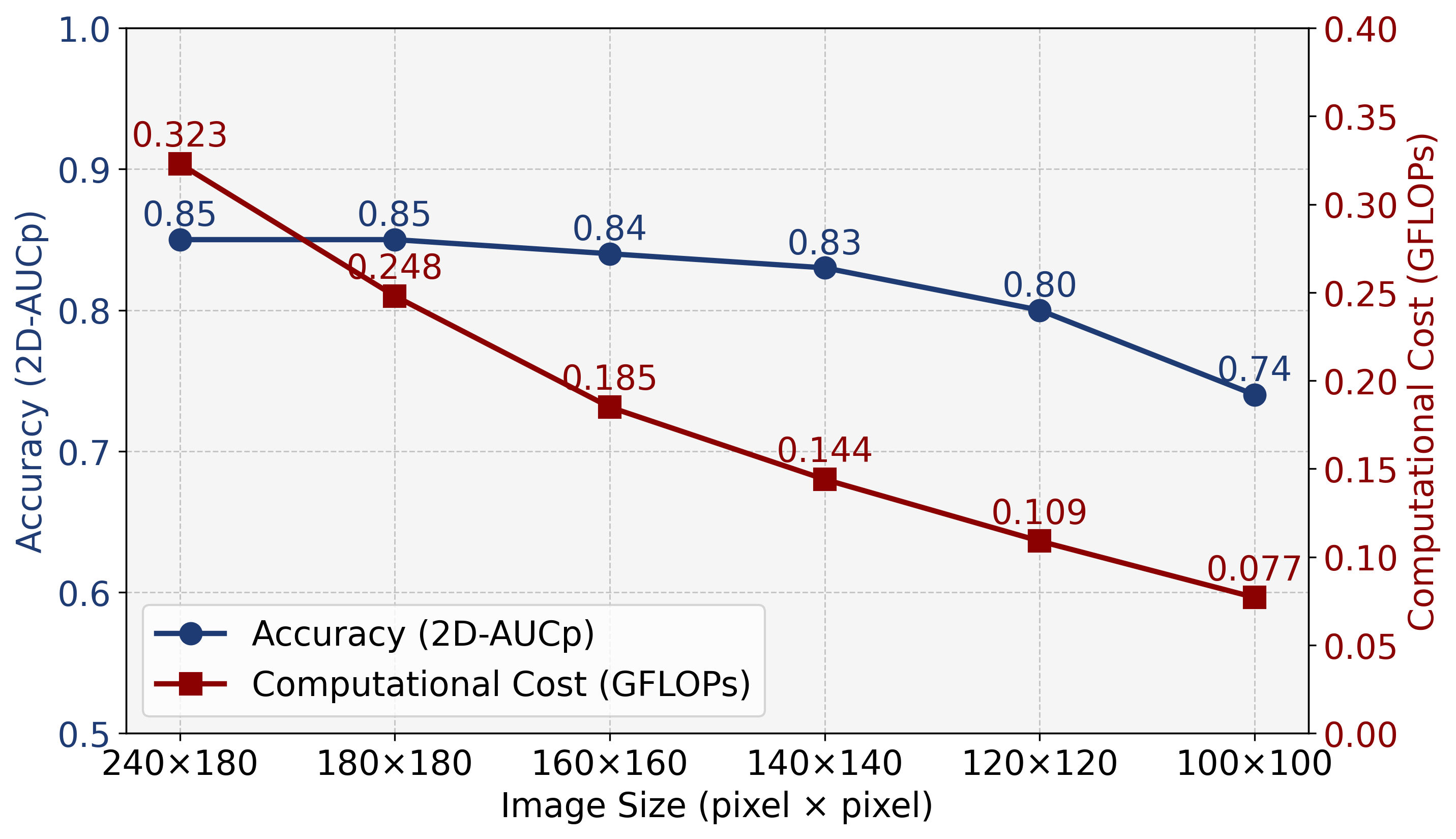}
    \caption{Impact of ROI size on accuracy (2D-AUCp) and computational cost (GFLOPs). Reducing the resolution substantially decreases FLOPs while maintaining competitive accuracy down to 160×160.}
    \label{Fig_ROISelection}
\end{figure}

Fig.~\ref{Fig_ROISelection} illustrates the variations in both performance and computational cost of our method under different ROI sizes. With only a minor accuracy drop of 0.01 (from 0.85 to 0.84), using an ROI size of 160×160 results in a 42.72\% reduction in computational cost. 
Therefore, we adopt this size for our method.

We perform comparative experiments on the MobileViT-V2 0.5× backbone to examine the effects of ROI and our auxiliary task. As shown in Table~\ref{Tab_ComparisonROIMTL}, adding the auxiliary task improves performance both with and without ROI, while ROI significantly reduces computational cost with slight accuracy drops. The auxiliary branch is pruned during inference, so its parameters and computation are not counted in the deployment cost. The combination of ROI and the auxiliary task is essential for achieving both high accuracy and low computational cost.

\begin{table}[t]
\centering
\caption{Ablation study on the wrist-based ROI and multi-task learning.}
\label{Tab_ComparisonROIMTL}
\begin{tabular}{cccc}
\toprule
\textbf{ROI} & \textbf{Multi-Task} & \textbf{2D-AUCp $\uparrow$} & \textbf{FLOPs $\downarrow$} \\
\midrule
No           & No                  & \textit{0.85}               & \textit{0.322 G}            \\
Yes          & No                  & 0.84                        & \textbf{0.185 G}            \\
No           & Yes                 & \textbf{0.87}               & \textit{0.322 G}            \\
Yes          & Yes                 & \textit{0.85}               & \textbf{0.185 G}            \\
\bottomrule
\end{tabular}
\end{table}

\section{Conclusion}
\label{Sect_Conclusion}

In this work, we presented EvHand-FPV, an efficient and lightweight framework for 3D hand tracking from a first-person view using event cameras. By introducing the wrist-based ROI detection, the LNES-Fast representation, and an optimized MobileViT V2-based multi-task learning network architecture, our method achieves 0.85 2D-AUCp on real FPV test set while reducing both parameter counts and computational cost by 89\% compared to EventHands. 
These results demonstrate that accurate, real-time egocentric hand tracking is feasible under stringent resource constraints, making our framework highly suitable for AR/VR headsets and other wearable platforms. 



\bibliographystyle{IEEEtran}
\bibliography{EHFPV}

\end{document}